\documentclass[a4paper, 10pt, conference]{ieeeconf}   

\IEEEoverridecommandlockouts\pubid{\makebox[\columnwidth]{979-8-3503-4544-5/23/\$31.00~\copyright{}2023 IEEE\hfill}\hspace{\columnsep}\makebox[\columnwidth]{ }}

\usepackage{graphicx}

\usepackage{tikz}
\usepackage{comment}
\usepackage{amsmath,amssymb} 
\usepackage{color}
\usepackage{multirow}
\usepackage{float}

\usepackage{epsfig}
\usepackage{subfig}
\usepackage[ruled,vlined]{algorithm2e}
\DeclareMathOperator{\rank}{rank}

\newcommand{\etal}{\textit{et al}. }
\newcommand{\ie}{\textit{i}.\textit{e}. }
\newcommand{\eg}{\textit{e}.\textit{g}.\ }

\usepackage[accsupp]{axessibility}  

\usepackage{FG2023}

\FGfinalcopy 

\IEEEoverridecommandlockouts                              
\def\FGPaperID{68} 

\title{\LARGE \bf
Laplacian ICP for Progressive Registration of 3D Human Head Meshes 
}

\author{\parbox{16cm}{\centering
    {\large Nick Pears$^1$, Hang Dai$^2$, Will Smith$^1$ and Hao Sun$^1$}\\
    {\normalsize
    $^1$ Department of Computer Science, University of York, UK\\
    $^2$ Mohamed bin Zayed University of Artificial Intelligence (MBZUAI), UAE }}
}

\begin{document}

\ifFGfinal
\thispagestyle{empty}
\pagestyle{empty}
\else
\author{Anonymous FG2023 submission\\ Paper ID \FGPaperID \\}
\pagestyle{plain}
\fi
\maketitle

\begin{abstract}

We present a progressive 3D registration framework that is a highly-efficient variant of classical non-rigid Iterative Closest Points (N-ICP).  Since it uses the Laplace-Beltrami operator for deformation regularisation, we view the overall process as Laplacian ICP (L-ICP). This exploits a `small deformation per iteration' assumption and is progressively coarse-to-fine, employing an increasingly flexible deformation model, an increasing number of correspondence sets, and increasingly sophisticated correspondence estimation. Correspondence matching is only permitted within predefined vertex subsets derived from domain-specific feature extractors. Additionally, we present a new benchmark and a pair of evaluation metrics for 3D non-rigid registration, based on annotation transfer. We use this to evaluate our framework on a publicly-available dataset of 3D human head scans (Headspace). The method is robust and only requires a small fraction of the computation time compared to the most popular classical approach, yet has comparable registration performance.

\end{abstract}

\section{Introduction}

\begin{figure}[!h]
\begin{center}
\begin{tabular}{c c}
\includegraphics[width=0.35\linewidth]{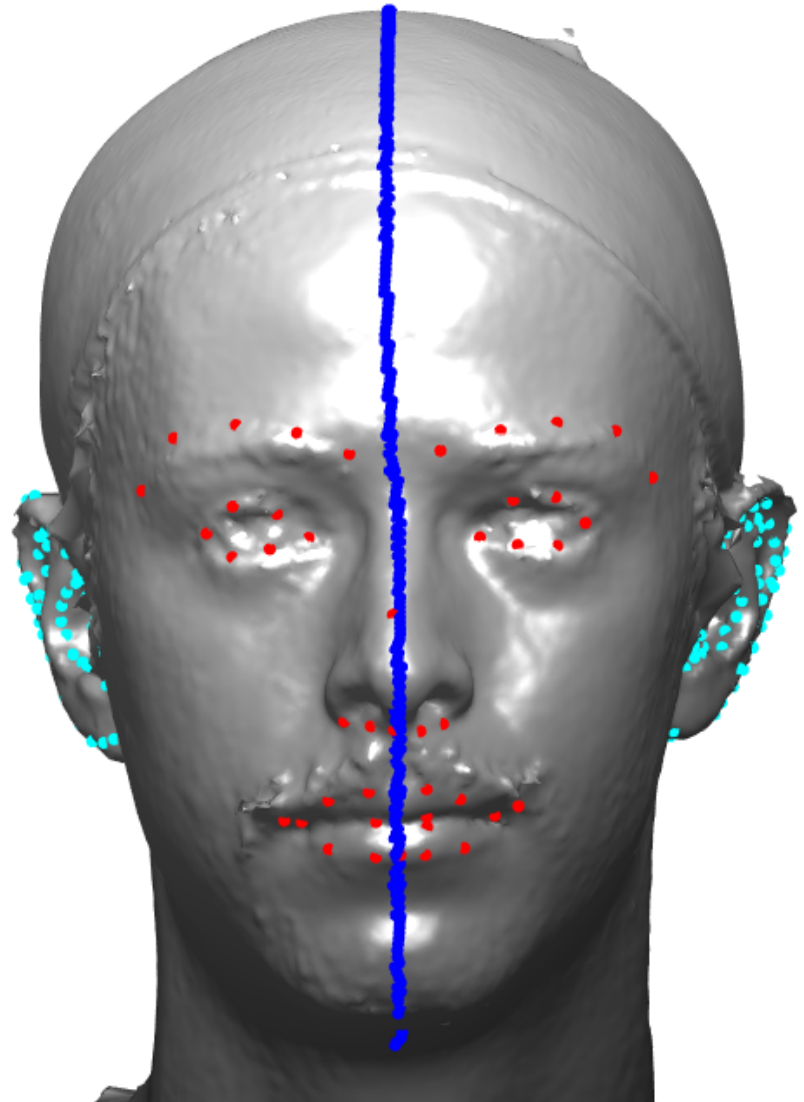}
&
\includegraphics[width=0.365\linewidth]{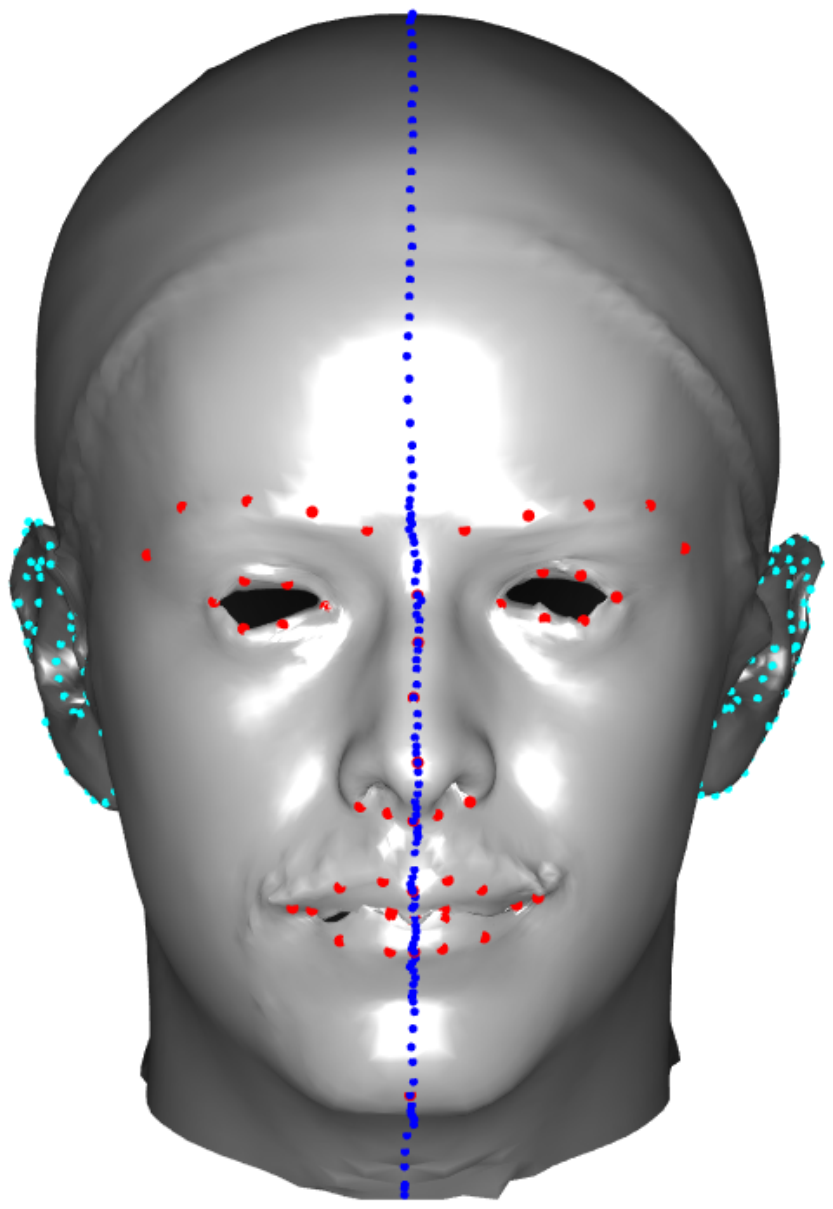} \\
Target Data  & Template Morph \\
\end{tabular}
\end{center}
\caption{Target scan (left) and template (right) morphed with Laplacian ICP. Correspondence sets are: (i) landmarks (red); (ii) right ear landmarks (cyan); (iii) left ear landmarks (cyan); (iv) symmetry contour (blue); and (v) all remaining vertices on mesh (grey surface).}
\label{fig:morphTeaser}
\end{figure}

\noindent
Determining surface correspondences across a set of 3D shapes is key to modelling them.
One approach employs non-rigid transformation of a template (source) shape, so that its vertices align with those of a target shape - see Figure \ref{fig:morphTeaser}. 
When a template is non-rigidly registered to a set of shapes of some class, 
this enables construction of statistical shape models, such as 3D Morphable Models (3DMMs, \cite{Amberg2007,ploumpis2020towards}).



Non-rigid registration has been extensively explored both in terms of classical optimisation algorithms and deep learning. Often, the latter requires a large corpus of training data, data augmentation techniques or transfer learning. Here, we revisit the classical approaches, which do not have such requirements and are of high utility in low data volume cases. In this respect, we provide a new formulation of dense, non-rigid Iterative Closest Points.

Our algorithm incorporates a form of progressively-relaxed Laplacian deformation regularisation into an progressively coarse-to-fine ICP-style framework  \cite{Besl1992,Amberg2007} - and hence our approach is termed Laplacian ICP (L-ICP). Laplacian mesh editing \cite{Sorkine2004} is known to be computationally efficient, due to its sparse linear structure and hence a key benefit of L-ICP is that it is very efficient computationally, when compared with competing classical approaches for non-rigid registration, such as Optimal-step N-ICP \cite{Amberg2007}. 
The approach exploits constraints supplied by domain-specific feature extraction algorithms, as shown in Figure \ref{fig:morphTeaser}, which has facial landmarks, ear landmarks and an intrinsic symmetry contour on a human head. 

L-ICP is implemented as a \emph{staged} non-rigid registration, where stages are defined by high-level scripting. Therefore, it is easily adapted to different shape classes, guided by domain-specific sets of extracted correspondences. It is a progressively coarse-to-fine  process that, as it transitions into each new stage, may employ: i) an increasing number of correspondence sets (landmarks, contours, regions); ii) an increasingly refined correspondence estimation; iii) an increasingly flexible shape deformation model and iv) an iterative shape refinement, used after correspondences stabilise, that generates a Laplace-Beltrami operator consistent with the updated template shape. 
We evaluate our work on the Headspace dataset of 3D human head scans \cite{Dai2019ijcv} 
and compare to the per-vertex affine regularisation approach from the most commonly-used N-ICP variant \cite{Amberg2007}.

Contributions are: i) fast and fully-dense morphing via progressively relaxed Laplace-Beltrami regularisation; ii) a flexible and progressive coarse-to-fine registration framework; iii) a new publicly-available non-rigid registration benchmark for the Headspace \cite{Dai2019ijcv} human head dataset, comprised of a set of manual annotations (859 subjects) and a pair of annotation transfer metrics. 
We will make our annotation data and registration code available in the interests of reproducibility.


\section{Related work}

Non-rigid 3D registration, correspondence matching and 3DMM fitting are highly-active research areas. Methods focus on organic shapes such as faces
 \cite{FLAME:SiggraphAsia2017,Dai2019ijcv,ploumpis2020towards}, human bodies \cite{SMPL:2015,Huang-NEURIPS2020,lombardi2021latenthuman} and various human organs \cite{koo2017} or man made objects, such as chairs, cups and aircraft \cite{Sung:2018,Liu-NIPS2020}.
A current popular approach is to employ implicit surface representations; for example, where the 3D surface is the zero level set of a learnt Signed Distance Function \cite{Park_2019_CVPR,Erler2020}.

In this work, we revisit classical ICP \cite{Besl1992} in its non-rigid form \cite{Amberg2007}.
Widely-used methods of classical non-rigid registration include Non-rigid Iterative Closest Point (N-ICP) \cite{Amberg2007}, Coherent Point Drift (CPD) \cite{Myronenko2010}, Thin Plate Spline (TPS) approaches \cite{bookstein1989,yang2011} and the method of Li \etal~\cite{Li2008}, which employs a Levenberg-Marquardt based optimisation.
The \emph{As-Rigid-As-Possible} (ARAP) form of  deformation regularisation was introduced by Sorkine \etal \cite{Sorkine2004} in their work on surface editing. 
Dai \etal \cite{Dai2018b} use the Laplace-Beltrami shape regularisation as a way of initialising the Coherent Point Drift \cite{Myronenko2010} algorithm.  Although we employ a similar early-stage template adaptation, \cite{Dai2018b} use fixed landmarks with a single fixed stiffness weight, whereas ours uses both fixed and variable correspondence sets and a stiffness weighting schedule. More importantly, we demonstrate, for the first time, that it is possible to use LB regularisation to do fast and dense shape morphing. 

\section{Laplacian ICP}


Inspired by Laplacian surface editing \cite{Sorkine2004}, we employ the Laplace-Beltrami (LB) operator in our shape regularisation term. When the LB operator is applied to a mesh, it extracts vectors in the direction of the local surface normal, with magnitude proportional to the local mean surface curvature. Here we employ the cotangent approximation scheme. Within our iterative optimisation scheme, we consider a regularising energy term, $E_{reg}$: 
\begin{equation}
E_{reg}({\bf X}_{i+1}) = \vert\vert {\bf L}_{i+1}{\bf X}_{i+1} -  {\bf L}_i{\bf X}_i \vert\vert_F ^2,
\label{eq:lb1}
\end{equation}
where 
${\bf X}\in\mathbb{R}^{N\times 3}$ is a matrix of $N$ source mesh vertex positions, ${\bf L}_i = L({\bf X}_i)$ is the LB operator (${\bf L}\in\mathbb{R}^{N\times N}$, sparse) computed from the source shape (\eg template) at the $i$th iteration and $\vert\vert.\vert\vert_F$ is the Frobenius norm.
This regularisation simultaneously applies an orientation constraint on the template update, because of the extracted surface normal directions - and a shape constraint, due to the surface normal magnitudes being proportional to local mean surface curvature. 
Note that $\rank({\bf L}) = N-1$. A physical interpretation of this is that a pure translation applied to all vertices of ${\bf X}$ would provide no change in the energy described by Eqn. \ref{eq:lb1}. The positional error associated with at least one pair of corresponding vertices from template to data can provide the necessary additional constraint as a shape error, $E_{shp}$, that we aim to minimise. We form a weighted combination of our energy terms as:
\begin{equation}
E = E_{shp} + \lambda E_{reg},
\end{equation}
where the parameter $\lambda$ balances the influence of the two component energies. Specifically, the energy for a new deformation is given as:
\begin{equation}
E({\bf X}_{i+1}) = \vert\vert {\bf P}_i {\bf X}_{i+1} -  {\bf Q}_i{\bf Y}_i \vert\vert_F ^2 + \lambda_i E_{reg}({\bf X}_{i+1}) 
\label{eq:lb2}
\end{equation}
where  ${\mathbf Y}_i$ are the target data vertices,  ${\mathbf P}_i$ and ${\mathbf Q}_i$ are highly-structured binary selection matrices that define source-target bijective correspondences and $\lambda_i$ is a weighting that defines the amount of mesh deformation regularisation. 


Clearly, Eqn. \ref{eq:lb2} is not closed form, as the regularisation term, $E_{reg}(.)$, at the update step $(i+1)$, is dependent on ${\bf L}_{i+1}$, which is a function of ${\bf X}_{i+1}$, \ie the updated template itself. This suggests an iterative procedure, where we invoke a \emph{small deformation} per-iteration assumption, so that
${\bf X}_{i+1} \approx {\bf X}_i$, which implies
${\bf L}_{i+1} \approx {\bf L}_i$ and that we can compute an accurate regularising term, $E_{reg}$. We ensure this small deformation assumption holds by initialising L-ICP with a high value of 
$\lambda_i$ giving a large template mesh `stiffness'. This parameter gradually becomes smaller as the template gets closer to the target data, thus balancing it with a smaller $E_{shp}$ term. 
In other words, we  employ a regularising mesh stiffness schedule, of gradually decreasing stiffness. 
This is analagous to the mesh stiffness schedule defined in N-ICP \cite{Amberg2007} that regulates mesh deformation by limiting differences in locally-affine transformations. 
Thus $\lambda$ changes in each iteration of L-ICP, as do the data vertices $\mathbf{Y}$, as any rigid component of template-data alignment is applied to the data rather than the template, for reasons described later.

We use a set of domain-specific feature extractors to generate $C$ subsets of source vertices (selected by ${\bf P}^{1 \dots C}$) and target vertices  (selected by ${\bf Q}^{1 \dots C}$) that are in correspondence with each other. Using these correspondence sets and the small deformation approximation for regularisation, we have our overall energy term as:
\begin{equation}
\small
E_{i+1} = \sum_{j=1}^{C} \alpha_j \vert\vert {\bf P}_i^j{\bf P}^j {\bf X}_{i+1} -  {\bf Q}_i^j{\bf Q}^j{\bf Y}_i \vert\vert_F ^2 
+
\lambda_i \vert\vert {\bf L}_i ({\bf X}_{i+1} -  {\bf X}_i) \vert\vert _F ^2 ,  
\label{eq:lb4}
\end{equation}
where $({\bf P}_i^j,{\bf Q}_i^j)$ select correspondences from within correspondence sets $({\bf P}^j,{\bf Q}^j)$. Note that $E_{i+1}=E({\bf X}_{i+1})$ and $\alpha_j$ is a weighting that expresses the relative confidence in the $j$th correspondence set. Eqn. \ref{eq:lb4} requires a linear solve for $3N$ variables per iteration, whereas N-ICP \cite{Amberg2007} requires a linear solve for $12N$ variables as per-vertex affine transformations are computed. Furthermore, the L-ICP constraint is vertex-based, whereas the N-ICP constraint is edge-based, giving around three times as many shape regulation equations in the linear solve for a triangular mesh. As a result, L-ICP is much more compact and efficient than N-ICP.


Suppose that, within some iteration, $i$, of L-ICP, we fix both the correspondences (all $C$ $\mathbf{P}_i,\mathbf{Q}_i$ matrices) and the parameter $\lambda_i$ and iteratively minimise the energy defined in Eqn. \ref{eq:lb4}  by updating the template mesh ${\bf X}$ (\ie employ an inner optimisation loop). This drives the regularisation term, $E_{reg}$, to zero, allowing the template to move closer to the data in steps of decreasing size until the recomputed template shape, ${\bf X}_{i+1}$, and the LB operator employed in the update, ${\bf L}_i$, become consistent with each other. In practice, we find that this second-order template deformation only takes a few iterations until the change in ${\bf X}$ over an iteration becomes small.  
This process provides a small template shape refinement and so is only used in the final morphing stage.

\section{Coarse-to-fine L-ICP framework}

\begin{figure*}[!h]
\begin{center}
\includegraphics[width= 0.6\linewidth]{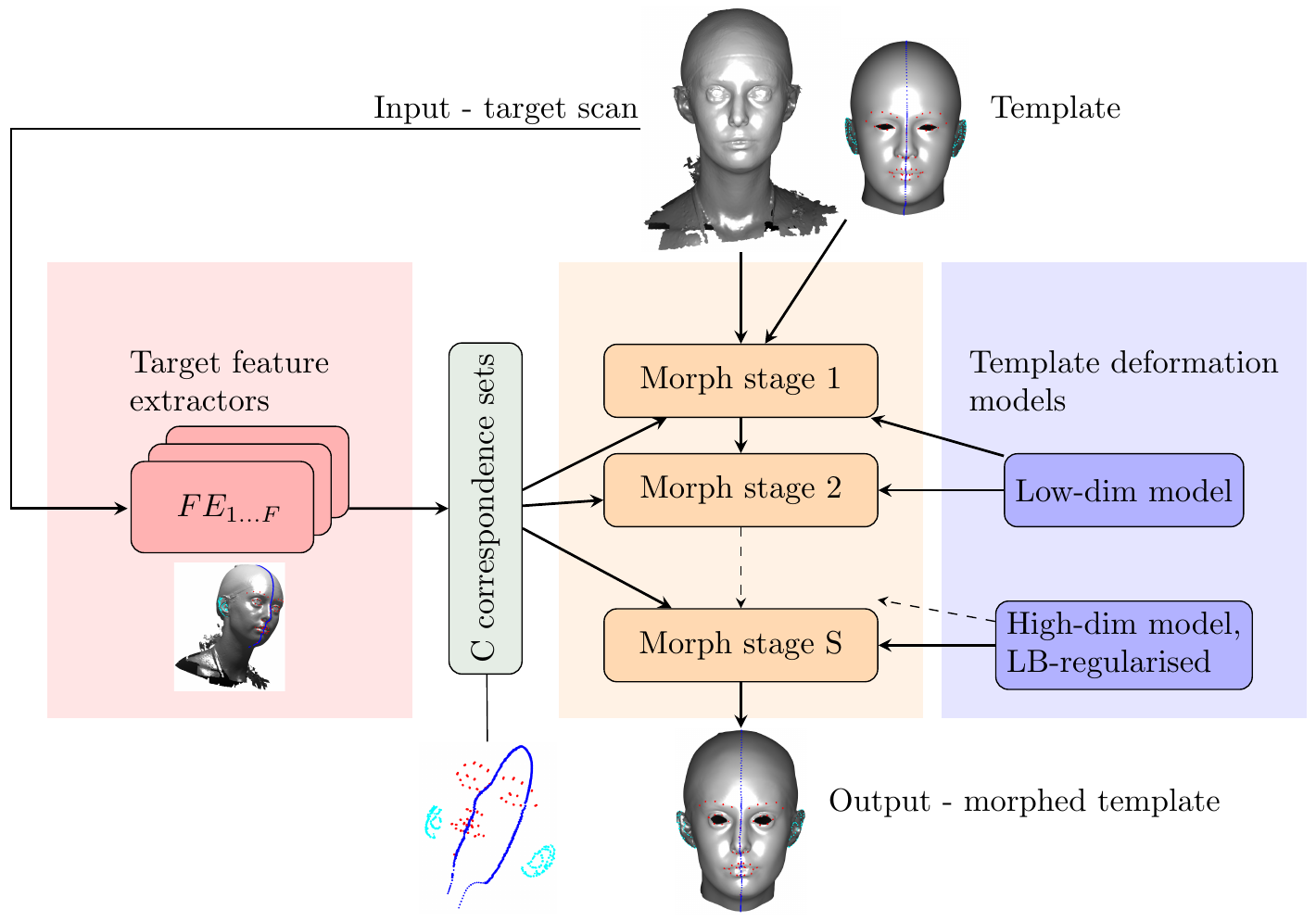} 
\end{center}
   \caption{L-ICP registration framework. Central orange panel: coarse-to-fine registration stages. Left pink panel: feature extractors for target data. Right purple panel: template deformation models. }
\label{fig:c2f}
\end{figure*}

Our L-ICP framework is shown in Fig. \ref{fig:c2f}. This defines a set of $s = 1 \dots S$ user-defined, application-dependent, coarse-to-fine registration stages, each of which terminates when the stage's template deformation falls below some threshold, $\vert\vert \Delta\mathbf{X} \vert\vert ^2_F < t_s$, or a maximum number of iterations, $i_s^{max}$ is reached. For each stage, the user defines: i) a set of correspondence sets; ii) a correspondence strategy; iii) a template deformation model, and iv) stage termination conditions, $(t_s, i_s^{max})$. The end user can rapidly define a morphing process as a set of such stages, as each stage inherits properties from the previous stage, unless they are re-specified. 
Early registration stages have a few landmark-based correspondence sets, coarse correspondence matching and low-dimensional deformation models (\eg global affine or low-dimensional 3DMMs). Later stages have many correspondence sets, fine correspondence matching and high-dimensional deformation models (\eg free movement with $3N$ degrees of freedom). 
A key feature is that we switch to a higher-dimensional deformation model before switching to dense correspondence search over the full surface. With otherwise free movement of the template vertices, this relies on the aforementioned Laplacian shape regularisation to effectively interpolate between the sparse correspondences, thus adapting the source template to the target shape. In essence, such \emph{template adaptation} is akin to a  fully-automatic, iterative mesh editing process and is the reason that our method is successful on widely different target shapes using a single template (\eg small babies heads and large adult heads).
Our framework exploits landmarks and/or contours and/or surface regions (\eg from semantic parts segmentation) that correspond across the source and target shapes. In the case of landmarks, a bijective (one-to-one) correspondence is predefined, and for contours and regions, we employ mutual nearest neighbour search.
We now detail the stage-selectable choices in our L-ICP framework that relate to Fig \ref {fig:c2f}.

\subsection{Correspondence sets for the human head}
\label{sec:corrsets}

We use L-ICP to register the FaceWarehouse head template \cite{Cao2014} with the Headspace dataset of 3D human heads \cite{Dai2019ijcv}. 
For the correspondence selection matrices, $\mathbf{P}_i^j,\mathbf{Q}_i^j$ in Eq. \ref{eq:lb4}, we employ a 3D face landmarker system, a 3D ear landmarker system, our own symmetry contour extractor, and a large correspondence set region that is all vertices that are otherwise unused in these landmark and contour sets. The face and ear landmark extractors employ the 2D channel as well as the 3D, whereas the symmetry contour extraction is based on 3D data only.

We use the standard {\tt dlib} face landmarker \cite{Kazemi2014}. This extracts 68 2D facial landmarks of which we retain 52, discarding the 16 that follow the apparent contour of the face. 
We project these 2D facial landmarks to their nearest vertices on the target 3D data scan.

The \emph{Human Ear Reconstruction Autoencoder} (HERA) system \cite{Sun2020} generates 55 landmarks per ear. This regresses the pose and shape parameters of a 3D Morphable Model (3DMM) of the human ear \cite{Dai2018}, such that a synthesised 2D image of the ear matches a rendered image of the colour-textured 3D target data. Two side views of the target 3D head can be rendered using facial landmarks, and the left ear is reflected enabling us to use a single right ear model. Again, 2D landmarks are projected to their nearest 3D vertices, and the residual Euclidean distance is stored, allowing the weighting of individual ear landmarks (larger distances have lower weights).

We adapt the method of Benz \etal \cite{Benz2002} for symmetry plane extraction into a more general procedure for symmetry contour generation. The piecewise nature of our algorithm allows the extracted symmetry contour to be intrinsic; for example, if the nose is bent to one side, it successfully tracks the nose ridge.

Finally the template/data vertices, $\mathbb{B}_{t,d}$, not designated as face/ear landmarks and not on the symmetry contour are defined for region-based correspondences 
\eg for the template: $\mathbb{B}_t = \{ \mathbb{X} \setminus \{ \mathbb{F}_t \cup \mathbb{E}^l_t \cup \mathbb{E}^r_t \cup \mathbb{S}_t \} \} $, where $\mathbb{X}$ is the set of all vertices on the templates and 
$\mathbb{F}_t, \mathbb{E}^l_t, \mathbb{E}^r_t, \mathbb{S}_t $ are the sets of template face landmarks, left ear landmarks, right ear landmarks and symmetry contour on the template respectively.


\subsection{Correspondence matching}
\label{sec:corrsearch}

We match across all active correspondence sets that do not consist of fixed landmarks. 
Our framework allows coarse and fine approaches to be selected.
Both are reliant on \emph{Mutual Nearest Neighbours} (MNN), where we take the subset of the bidirectional 1-nearest neighbour search results, such that the correspondence relation is bijective. Additionally, this mutual nearest neighbour search can specified to be over the 3-DOF vertex positions or 6-DOF vertex positions and their associated normal vectors, with an weighting factor between positions and normals.
Firstly, MNN search in itself is a suitably conservative approach for early-stage morphing and has the benefits of handling mesh holes automatically and obviating the need for a manually-tuned threshold to filter out bad correspondences. 
Secondly, in our \emph{normal shooting} method, the MNN approach is augmented by projecting a vector from the template to its corresponding data vertex along the source normal, which in general, results in an off-surface target point. However, when the source and target are close to each other, this is often a better morph direction, due to the generally different spatial sampling of the two surfaces.  

\subsection{Deformation models}
\label{sec:defmodels}
Early stages of deformation use a global affine model and later stages use LB-regularised template deformation.

\subsubsection{Global affine deformation}
\label{sec:affinedef}

Suppose that the template shape is given as a matrix of vertex positions, ${\bf X_i}\in\mathbb{R}^{N \times 3}$, after the $(i-1)th$ shape update with initial shape $\mathbf{X}_1$, and that the data, whose 6 DoF pose may vary, is given as the matrix of vertex positions  ${\bf Y_i}\in\mathbb{R}^{N_Y\times 3}$, then we solve the following linear least squares equation for the affine transform $\mathbf{A_i}\in\mathbb{R}^{4\times 3}$:

\begin{equation}
 \begin{bmatrix}  \alpha_1\mathbf{P}^1_{i} \mathbf{P}^1 \mathbf{X_i} ~~\alpha_1\mathbf{1}^1_{i}  \\
                    \vdots    ~~~~~~~~~~~~~~ \vdots                         \\
                  \alpha_C\mathbf{P}^{C}_{i} \mathbf{P}^C \mathbf{X_i} ~~\alpha_C\mathbf{1}^C_{i} 
 \end{bmatrix} \mathbf{A}_i = 
 \begin{bmatrix} 
  \alpha_1\mathbf{Q}^1_{i} \mathbf{Q}^1 \mathbf{Y_i} \\
    \vdots  \\
   \alpha_C\mathbf{Q}^{C}_{i} \mathbf{Q}^C \mathbf{Y_i} 
 \end{bmatrix},
 \label{eqn:affineComp}
\end{equation}
where $(\alpha_1 \dots \alpha_C)$  are the relative influence weights for various sets of correspondences.
$\mathbf{P}_{i}^j \in\{0,1\}^{N^j\times N^j_a}$ is a binary selection matrix that selects $N^j$ vertices associated with the $j$\emph{'th} correspondence set ($j= 1 \dots C)$ from the set of \emph{all} $N^j_a$ vertices in that correspondence set and $\mathbf{P}^j \in\{0,1\}^{N^j_a\times N}$ is the binary matrix that selects \emph{all} members of the correspondence set from the template vertices. 
Note that if the correspondence set contains fixed landmarks, then 
$\mathbf{P}_{i}^j = \mathbf{I}_{N^j_a}$, otherwise it is determined by mutual nearest neighbour correspondence search, whereas $\mathbf{P}^j$ are constant matrices. 
Also note that $\mathbf{1}_i^j$ is a vector of 1s with length equal to the number of correspondences on iteration $i$ for correspondence set $j$.
The matrices $\mathbf{Q}_{i}^j, \mathbf{Q}^j$ are binary selection matrices that select data vertices in an analagous way to the template selection matrices. 
We choose to decompose the affine transform into a rigid part and a non-rigid part, such that
\begin{equation}
\mathbf{A}_i = \begin{bmatrix} \mathbf{R}_i\mathbf{B}_i \\ \mathbf{t}_i \end{bmatrix}
\label{eqn:Adecomp}
\end{equation}
where $\mathbf{R}_i\in\mathbb{R}^{3\times 3}$ is a rotation matrix, $\mathbf{B}_i\in\mathbb{R}^{3\times 3}$ is a non-rigid deformation matrix composed of anisotropic scaling and shears and $\mathbf{t}_i\in\mathbb{R}^{1\times 3}$ is a translation vector. We then apply the rigid part of the affine transform to the data:
\begin{equation}
\mathbf{Y}_{i+1} = (\mathbf{Y}_i - \mathbf{1_{N_Y}}\mathbf{t}_i) \mathbf{R}_i^\mathrm{T} 
\label{eqn:Yupdate}
\end{equation}
where $\mathbf{1_{N_Y}}$ is a column vector of $N_Y$ 1s. The non-rigid part of the affine transform is applied to the template:
\begin{equation}
\mathbf{X}_{i+1} = \mathbf{X}_i\mathbf{B}_i    
\label{eqn:Xupdate}
\end{equation}
We could apply the full affine transform to the template, so this may seem like an unnecessary complication. However, it is very useful to employ this decomposition, which maintains a canonical pose of the template, particularly in variants of L-ICP that constrain template deformation to be symmetrical, or employ a 3DMM to reduce the dimensionality of the template deformation model. 

\subsubsection{LB regularised template deformation}
\label{sec:lbregdef}

In later stages of the morphing process, we wish to deform the template in a more detailed way that cannot be modelled by a simple low-dimensional transform. To achieve this, we solve directly for source mesh vertex positions, under the regularisation of the Laplace-Beltrami constraint. Specifically, we minimize the energy in Equation \ref{eq:lb4} by iteratively solving for $\mathbf{X}_{i+1}$ in the following weighted linear least-squares problem:
\begin{equation} 
 \begin{bmatrix}    
                    \alpha_1\mathbf{P}^1_{i} \mathbf{P}^1  \\
                    \vdots       \\
                     \alpha_C\mathbf{P}^{C}_{i} \mathbf{P}^C  \\
                   \lambda_i\mathbf{L_i} 
 \end{bmatrix}      \mathbf{X}_{i+1} =  \begin{bmatrix}
                    \alpha_1\mathbf{Q}^1_{i} \mathbf{Q}^1 \mathbf{Y}_i \\
                   \vdots  \\
                \alpha_C\mathbf{Q}^{C}_{i} \mathbf{Q}^C \mathbf{Y}_i \\
                \lambda_i\mathbf{L_i}\mathbf{X}_i 
\end{bmatrix},
\label{eq:LTA}
\end{equation}
 where $\alpha_j$ are relative influence weights for various sets of correspondences and $\lambda_i$ is the mesh stiffness weight at iteration $i$ of the deformation stage.

\section{3D registration of the human head}
\label{sec:headreg}
\begin{figure*}[!ht]
\begin{tabular} {c}
\includegraphics[width=0.9\linewidth]{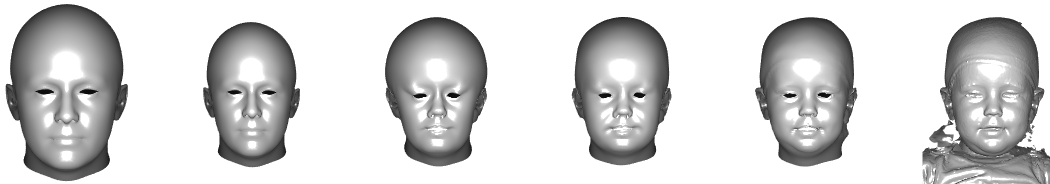} \\ 
\end{tabular}
\caption{Far left: Facewarehouse \cite{Cao2014} head template. Far right: the raw 3D Headspace data \cite{Dai2019ijcv}. Intermediate images show morph stages 2-5 of our L-ICP framework. In most applications, only the final registration quality matters (column 5) and we see that the poor shape around the eyes after stage 3 (column 3) has been corrected. Zoom required.}
\label{fig:morphStages}
\end{figure*}

Correspondence sets are weighted using empirical grid search as follows: face landmarks 1.5, symmetry contour 1.4, left/right ear landmarks 1.0, all other vertices 1.0.
We define five stages for human head registration, with per-stage output examples shown in Fig. \ref{fig:morphStages}.

\noindent
{\bf Stage 1 - Affine template initialisation.}
\label{sec:affineinit}
Goal: align the data to the template, transform the template to the same scale and aspect ratio of the data. Settings: i) correspondence sets, $C=3$: face landmarks, left ear landmarks, right ear landmarks; ii) correspondence matching: MNN; iii) deformation model:  global affine (one shot).
We solve Eq \ref{eqn:affineComp} for the required global affine deformation, which is then decomposed and distributed between the target data (rigid part) and the template (non-rigid part), as described in Eq  \ref{eqn:Adecomp} to  \ref{eqn:Xupdate}.

\noindent
{\bf Stage 2 - Affine template adaptation.}
Goal: improve depth and height scaling using symmetry contour. Settings: i) correspondence sets, $C=4$ : three landmark sets from previous stage plus the symmetry contour; ii) correspondence matching: MNN; iii) deformation model: global affine (iterative). We iteratively compute the affine template update, $\mathbf{A}_i$ using Eq \ref{eqn:affineComp} and perform the template/target updates according to Eq \ref{eqn:Adecomp} to  \ref{eqn:Xupdate}. 
Maximum iterations is set at 15. 


\noindent
{\bf Stage 3 - Laplacian template adaptation.}
Goal: adapt the template shape to the landmarks and symmetry contour.
Settings: i) correspondence sets, $C=4$, same as previous stage; ii) correspondence matching: MNN; iii) deformation model: LB-regulated free vertex deformation (iterative).
The stiffness parameter, $\lambda_i$ ranges from 100 to 0.1 with a maximum number of iterations set at 58.


\noindent
{\bf Stage 4 - Morphing with dense correspondences.}
Goal: compute dense correspondences for dense morphing.
Settings: i) correspondence sets, $C=5$: all landmark sets plus symmetry contour plus the `set difference' region; ii) correspondence matching: MNN; iii) deformation model: LB-regulated free vertex deformation (iterative).
The stiffness parameter ranges from 100 to 1 with maximum iterations set to 31.

\noindent
{\bf Stage 5 - Morphing with normal shooting.}
Goal: employ more refined dense correspondences for dense morphing.
This stage re-specifies the correspondence search to the \emph{normal shooting} correspondence method, but otherwise inherits all other framework selections from stage 4. 
The stiffness parameter ranges from 0.9 to 0.1 with maximum iterations set to 27. This means the maximum number of shape change iterations is 132 over all stages.

\section{Evaluation}


Quantitative evaluation of 3D shape registration and correspondence quality using real-world data is notoriously difficult due to a lack of high-quality ground truth data. 
One approach is to use a proxy evaluation where better correspondences are deemed to be those that build better statistical models according to some metrics. For example, Styner \etal \cite{styner2003evaluation} proposed the use of three 3DMM metrics - compactness, generalisation and specificity. However, these are only meaningful when the template is on or very close to the data surface.
Furthermore, for soft organic shapes, strong perceptual consensus on what is a good correspondence often only exists at a very sparse set of surface locations. In our human head example, these are the physical junctions of tissues, such as eye and mouth corners. 
To mitigate these problems, we propose a different form of evaluation procedure and benchmark, which is based on the manual annotation of \emph{facial contours}. Quantitative metrics are proposed that capture both the repeatability and homogeneity of how such annotations are transferred from the data onto the non-rigidly registered template. These are detailed further in Section \ref{sec:anntrans}, but we first describe the dataset used.

\subsection{Dataset}

For evaluation, we use the publicly-available Headspace dataset \cite{Dai2019ijcv} of high resolution (150K-200K vertices) 3D images of the human head. 
We employ two of the five 2D views (left-frontal and right-frontal) to manually annotate a range of facial feature contours, including those around the eyes and eyelid, mouth, nose, nasolabial folds and ears. 
In the case of facial contours that are not well-defined in the image pair, annotators are instructed to omit them. Due to the labour-intensive nature of this, we have around 57\% coverage of the Headspace dataset.


\subsection{Annotation transfer}
\label{sec:anntrans}
\begin{figure}
\begin{center}
\begin{tabular}{c c | c c | c}
     
      
      \includegraphics[width=0.19\linewidth]{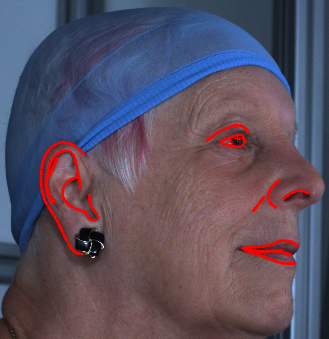} &
      \includegraphics[width=0.163\linewidth]{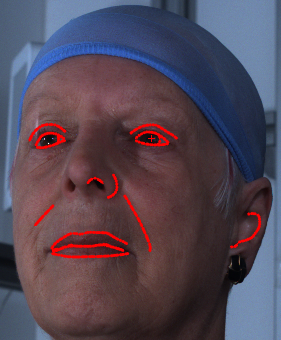} &
      \includegraphics[width=0.15\linewidth]{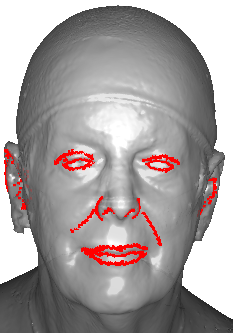} &
       \includegraphics[width=0.145\linewidth]{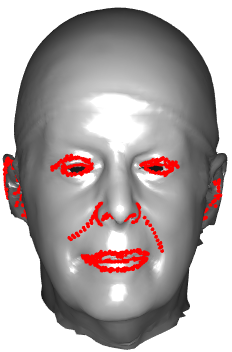}   &
        \includegraphics[width=0.14\linewidth]{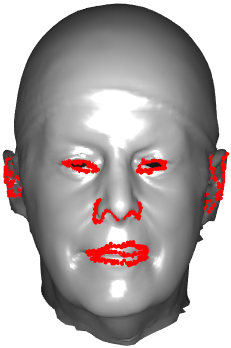} \\         
                
    \includegraphics[width=0.19\linewidth]{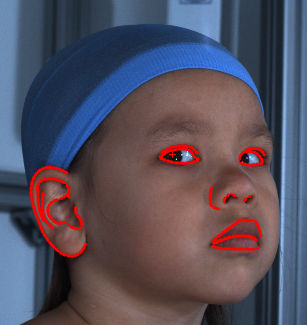} &
      \includegraphics[width=0.169\linewidth]{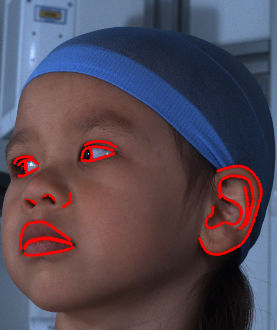} &  
    \includegraphics[width=0.15\linewidth]{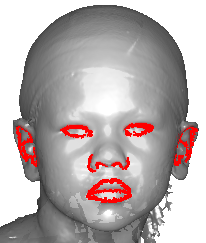} &
      \includegraphics[width=0.145\linewidth]{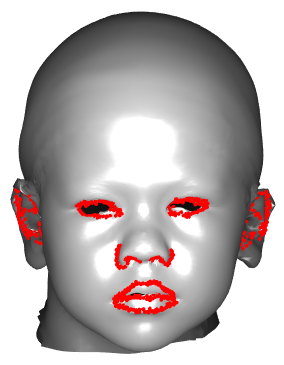} &
        \includegraphics[width=0.1375\linewidth]{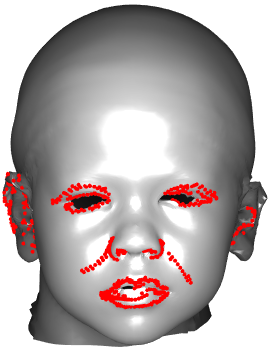}
  
\end{tabular}
\end{center}
\caption{Left two columns: manual annotations. Third column: annotations amalgamated and projected to 3D target scans. Fourth column - morphed templates with annotations transferred. Right column - annotations swapped across subjects.}
\label{fig:annotationTransfer}
\end{figure}

We transfer the 2D target data annotations to the 3D target mesh, which is achieved via left/right camera ray-to-mesh intersection, where rays are generated using the left/right camera calibration matrices. These 3D target scan mesh surface coordinates are then transferred on to their nearest neighbor template mesh vertices, after registration. If multiple annotations (\eg left and right view of the same facial contour) transfer to the same 3D template coordinate, that is recorded as a single intersection. Examples of annotation transfer over a wide subject age range are given in Figure \ref{fig:annotationTransfer}.

\begin{figure}
\begin{center}
\includegraphics[width=0.7\linewidth]{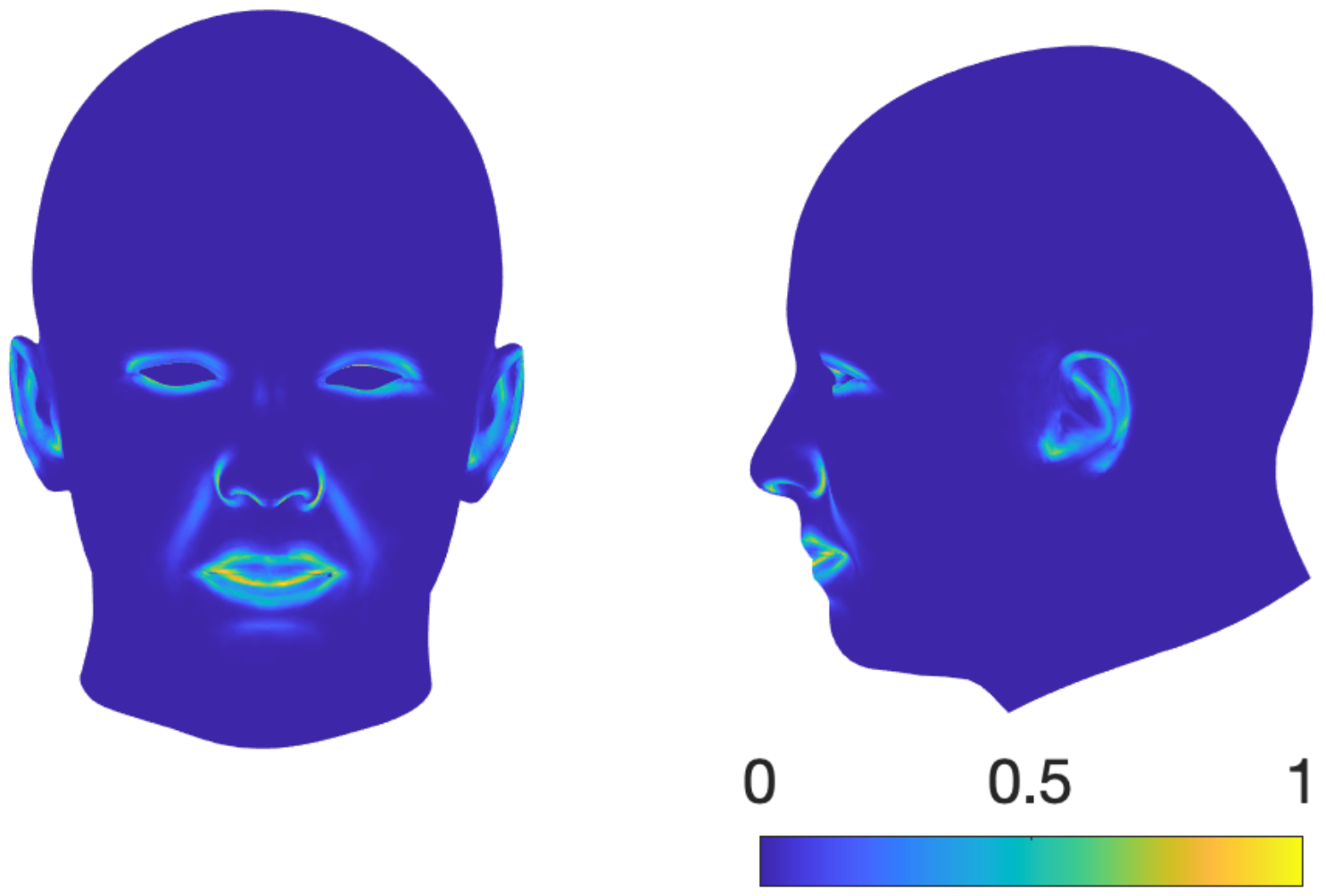}
\end{center}
\caption{Annotation transfer density color map for $N_{subj}=675$ Headspace subjects   \cite{Dai2019ijcv} after L-ICP registration. We omit 184 annotated subjects where there is not a full set of ear landmarks due to hair/cap occlusions.  
}
\label{fig:annotationTransferColormap}
\end{figure}

For consistent non-rigid registration, annotations on different subject's target data scans should transfer to the same vertices on the template, or at least to closely neighboring vertices. 
Thus, by measuring the template surface density of such an \emph{annotation transfer} process, we can generate a quantitative evaluation of registration repeatability.
Additionally, this can be qualitatively evaluated by color-mapping the template with the density of those annotation transfers. These should be sharply defined on the template. Figure \ref{fig:annotationTransferColormap}-left shows a front and side view of the employed Facewarehouse \cite{Cao2014} template, with its surface color mapped with the frequency of the annotation transfers over $N_{subj}=675$ Headspace subjects, with dark blue being zero transfers and yellow being the maximum frequency of transfer. The transfer is highly-repeatable around facial features, especially the mouth and nose. This is expected, as the registration process is well-guided by automatic landmarking in these regions. In contrast, there is lower repeatability around the nasolabial folds. 


\subsubsection{Annotation transfer metrics}
A metric should indicate high performance when template vertices are selected in the transfer numerous times (in the best case, $N_{subj}$ times) and low performance when they are selected few times (in the worst case, once). 
Let $v \in \mathbb{V}$ be the set of template vertex indices that have at least one annotation transfer ($t_v \ge 1)$ and denote the vertex set cardinality (for vertices with non-zero $t_v$) be  $\vert \mathbb{V} \vert$. We define the mean \emph{annotation transfer density} as:
\begin{equation}
\bar{d} = \frac{1}{N_{subj}\vert\mathbb{V}\vert} \sum_{v \in \mathbb{V}} t_v
\end{equation}
where $N_{subj}$ is the number of subject target scans employed in the evaluation.

The density metric is straightforward to apply and is a quantitative measure that relates directly to qualitative annotation density colormaps. 
However, it is annotation-label agnostic
and does not handle the case when annotation contours are in close proximity to each other (\eg bottom of upper lip and top of lower lip). Here, contours with different semantic labels may transfer to the same morphed template vertices. Ideally, template vertices are selected by annotations of a single semantic label. Therefore, we additionally define a mean \emph{annotation transfer homogeneity} metric. To do this, we  define $t_{v,i} \ge 1$ as the non-zero number of annotation transfers of type $i$, for vertex $v \in \mathbb{V}_i$, where $i$ indexes a semantic annotation label in the full set of annotation labels, $\mathbb{A}$. The mean homogeneity is then:
$\bar{h} =\sum_{i \in \mathbb{A}} \omega_i\bar{h}_i$, where
\begin{equation}
\bar{h}_i = \frac{\sum_{v \in \mathbb{V}_i}  t_{v,i}} {\sum_{v \in \mathbb{V}_i} \sum_{j \in \mathbb{A}} t_{v,j}},~~~ 
w_i = \frac{\sum_{v \in \mathbb{V}_i} t_{v,i} }{ \sum_{j \in \mathbb{A}}\sum_{v \in \mathbb{V}_j}t_{v,j}}
\end{equation}
Here, $\bar{h}_i$ is mean homogeneity per annotation label $i$ and $\omega_i$ is a weighting based on the relative prevalence of transfers of that annotation label, with $\sum_{i \in \mathbb{A}} \omega_i =1$.



A limitation of these metrics is that they are template mesh specific.
Templates of higher resolutions will give lower values for these metrics, since a given template vertex can only be selected for the annotation transfer (\ie as the 1-nearest neighbour) over a smaller surface area. It may be possible to use an additional normalising factor, $\alpha_{res}$, that adjusts for template resolution, but this may be confounded by non-uniform template resolutions. 

\begin{figure}
\begin{center}
\begin{tabular}{c c}
\includegraphics[width=0.45\linewidth]{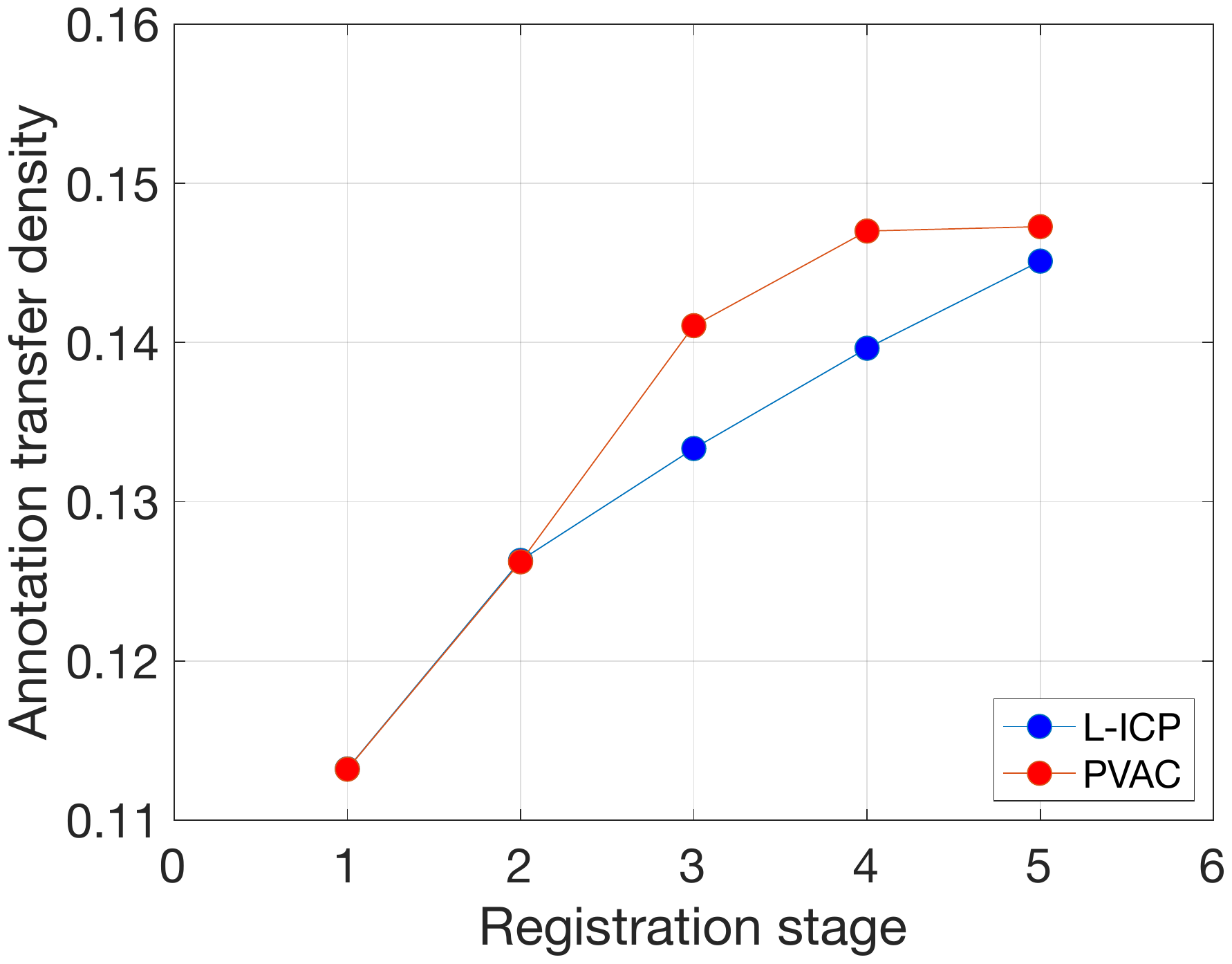} &
\includegraphics[width=0.45\linewidth]{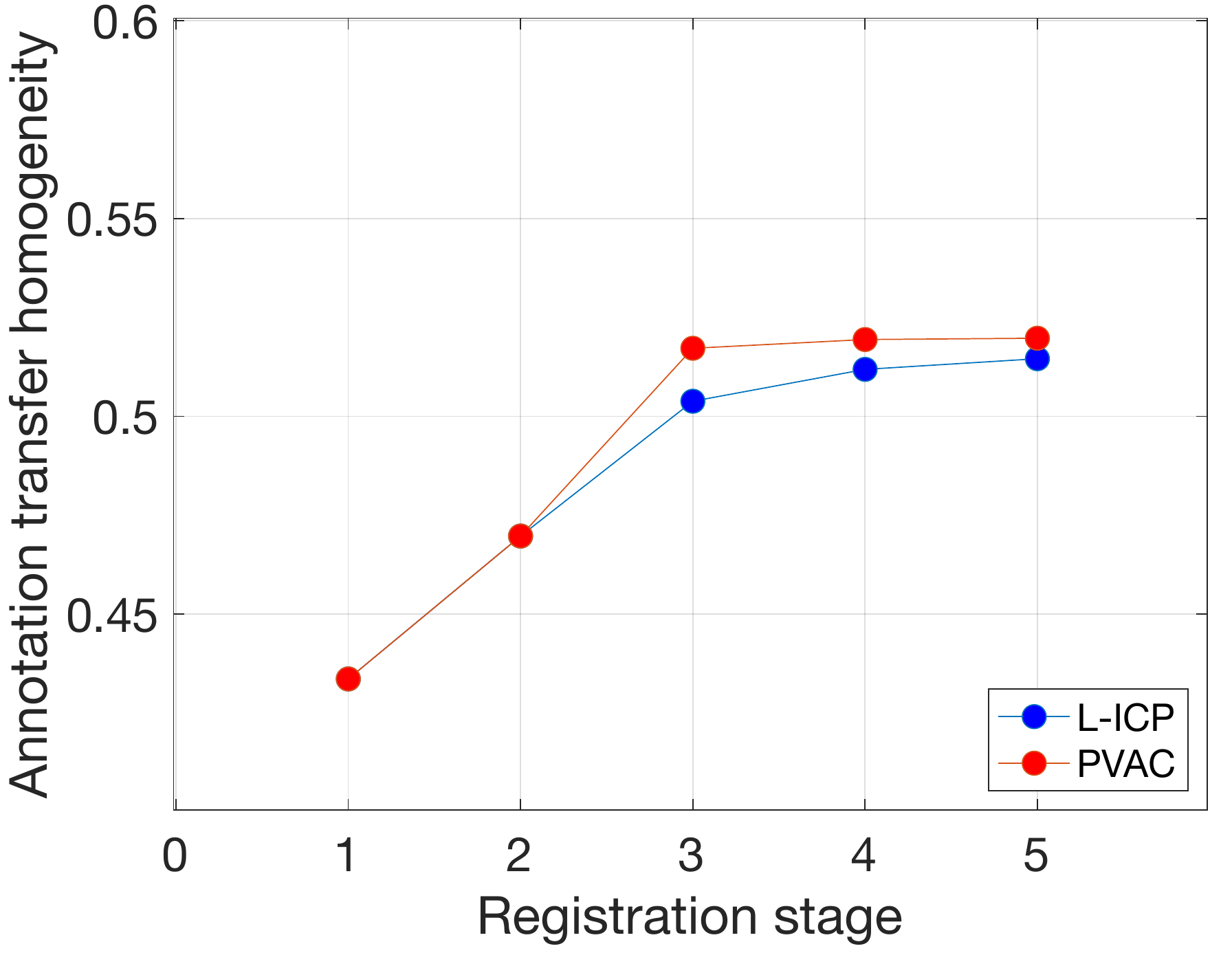}  \\
\multicolumn{2}{c} {\includegraphics[width=0.9\linewidth]{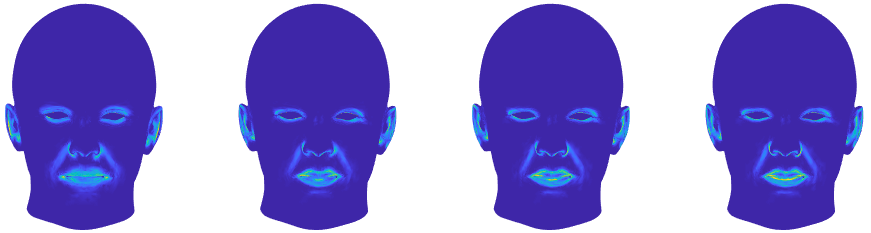}}
\end{tabular}
\end{center}
\caption{Top left: annotation transfer density, 118 subjects over five registration stages. PVAC employs the \emph{Per Vertex Affine Constraint} \cite{Amberg2007} within our framework. Top right: annotation transfer homogeneity. Bottom row: colormaps of annotation transfer density: L-ICP stages two (left) to five (right).}
\label{fig:metrics}
\end{figure}

In Figure \ref{fig:metrics} we compare L-ICP with a version of our framework that has the per-vertex affine constraint (PVAC) \cite{Amberg2007} substituted for Laplacian regularisation, with all parameters the same. We compare the two approaches in terms of annotation density and homogeneity over five stages of morphing. (This is done over 118 subjects due to high PVAC computation time.) Although the PVAC constraint approaches its maximum in fewer stages, the final performance is very similar at a fraction of the computational cost. 

\subsection{Processing time}

Table \ref{tab:times} gives the average processing times in seconds for L-ICP to reach the \emph{end} of each stage (averaged over 118 scans). Scans are typically 150K-200K vertices with a template of size 11.51K vertices. This was evaluated on a Macbook Pro with 2.3 GHz Quad-Core Intel Core i7, 32GB of memory, macOS Big Sur, running Matlab version R2021a. L-ICP is over 26 times faster at stage 5, than when a per-vertex affine constraint (PVAC) is employed within the same coarse-to-fine framework, with the same features and using the same stiffness schedule. Thus, L-ICP gives a dense, consumer laptop-based registration in around 47 seconds compared to around 20 minutes for PVAC.

\vskip .1cm
\begin{center}
\begin{table}[h]
\begin{tabular}{ |c||c|c|c|c|c| } \hline
 Stage           & 1 & 2 & 3 & 4 & 5 \\  \hline\hline
L-ICP (s)      & 0.03 & 0.28 & 15.68 & 32.38 & 47.13 \\  
PVAC (N-ICP) (s)       & 0.03 & 0.24 & 252.86 & 1183.67 & 1230.51  \\  \hline
\end{tabular}
\caption{Annotation transfer metrics for L-ICP and N-ICP}
\label{tab:times}
\end{table}
\end{center}
\vskip .1cm

\section{Conclusions}

 We have demonstrated that fully-automatic, \emph{densely-corresponded} non-rigid registration only requires a Laplacian for a regularisation term and hence is rapid to compute. It achieves this via a \emph{small deformation per iteration} assumption within a progressive coarse-to-fine framework that is guided by within-set correspondences from application-specific feature extractors. Registration performance is comparable with using per-vertex affine constraints in standard N-ICP \cite{Amberg2007}, but at considerably lower computational cost. Finally, we presented a new benchmark for registration based on contour sketch annotations and a pair of annotation transfer metrics.

%
%
\bibliographystyle{ieee}
\bibliography{egbib}

\end{document}